\documentclass[10pt,twocolumn,letterpaper]{article}

\usepackage[pagenumbers]{cvpr} % To force page numbers, e.g. for an arXiv version

\usepackage[dvipsnames]{xcolor}
\usepackage{algorithm}
\usepackage{algorithmic}
\usepackage{multirow}

\definecolor{cvprblue}{rgb}{0.21,0.49,0.74}
\usepackage[pagebackref,breaklinks,colorlinks,citecolor=cvprblue]{hyperref}
\def\paperID{10328} % *** Enter the Paper ID here
\def\confName{CVPR}
\def\confYear{2024}

\title{MAPSeg: Unified Unsupervised Domain Adaptation for Heterogeneous Medical Image Segmentation Based on 3D Masked Autoencoding and Pseudo-Labeling}

\author{Xuzhe Zhang$^{1 \dagger}$, Yuhao Wu$^{2 \dagger}$, Elsa Angelini$^{1,3}$, Ang Li$^{4}$, Jia Guo$^{1}$, Jerod M. Rasmussen$^{5}$, \\Thomas G. O'Connor$^{6}$, Pathik D. Wadhwa$^{5}$, Andrea Parolin Jackowski$^{7}$, \\Hai Li$^{2}$, Jonathan Posner$^{2}$, Andrew F. Laine$^{1 \ddagger}$, Yun Wang$^{2,8\ddagger}$ \\
$^1$Columbia University\quad$^2$Duke University\quad$^3$Télécom Paris, LTCI, Institut Polytechnique de Paris
\\$^4$University of Maryland, College Park\quad$^5$University of California, Irvine\quad$^6$University of Rochester\\$^7$Universidade Federal de São Paulo\quad$^8$Emory University}
\begin{document}
\maketitle
\def\thefootnote{$\dagger$}\footnotetext{Co-first authors.}\def\thefootnote{\arabic{footnote}}
\def\thefootnote{$\ddagger$}\footnotetext{Co-senior supervising authors.}\def\thefootnote{\arabic{footnote}}
\begin{abstract}
Robust segmentation is critical for deriving quantitative measures from large-scale, multi-center, and longitudinal medical scans. Manually annotating medical scans, however, is expensive and labor-intensive and may not always be available in every domain. Unsupervised domain adaptation (UDA) is a well-studied technique that alleviates this label-scarcity problem by leveraging available labels from another domain. In this study, we introduce Masked Autoencoding and Pseudo-Labeling Segmentation (MAPSeg), a \textbf{unified} UDA framework with great versatility and superior performance for heterogeneous and volumetric medical image segmentation. To the best of our knowledge, this is the first study that systematically reviews and develops a framework to tackle four different domain shifts in medical image segmentation. More importantly, MAPSeg is the first framework that can be applied to \textbf{centralized}, \textbf{federated}, and \textbf{test-time} UDA while maintaining comparable performance. We compare MAPSeg with previous state-of-the-art methods on a private infant brain MRI dataset and a public cardiac CT-MRI dataset, and MAPSeg outperforms others by a large margin (10.5 Dice improvement on the private MRI dataset and 5.7 on the public CT-MRI dataset). MAPSeg poses great practical value and can be applied to real-world problems. GitHub: \url{https://github.com/XuzheZ/MAPSeg/}.

\end{abstract}    
\section{Introduction}
\label{sec:intro}
\begin{figure}[t!]
  \centering
     \includegraphics[width=0.78
     \linewidth]{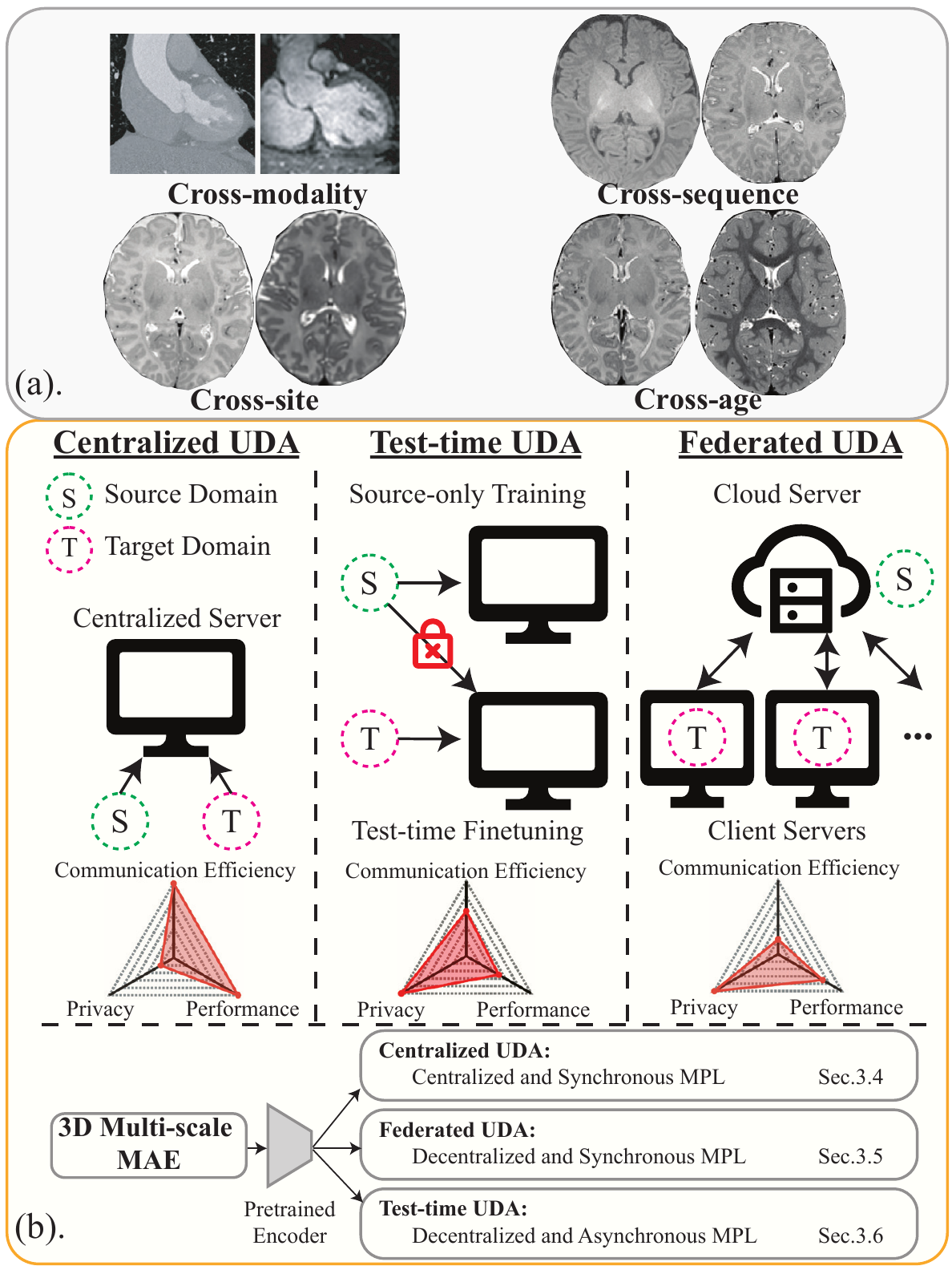}
  \caption{(a). Illustrations of four different domain shifts in medical images. (b). Overview of different UDA settings and how MAPSeg can fit into different scenarios.}
  \label{fig:overview}
  \vspace{-1.5em}
\end{figure}

Quantitative measures from medical scans serve as biomarkers for various types of medical research and clinical practice. For instance, neurodevelopmental studies utilize metrics such as brain volume and cortex thickness/surface area from infant brain magnetic resonance imaging (MRI) to investigate the early brain development and neurodevelopmental disorders~\cite{cruz2023cortical, shen2022subcortical, baribeau2019structural, hazlett2017early}. Therefore, robust segmentation of medical images acquired from large-scale, multi-center, and longitudinal studies is desired, yet often challenged by the domain shifts across different imaging techniques and even within a single modality (\hyperref[fig:overview]{Fig.1a}). For example, computed tomography (CT) and MRI provide markedly different signals for the same structure (\eg, cardiac regions, \hyperref[fig:overview]{Fig.1a}). MRI, a widely adopted radiation-free imaging technique, bears various types of inherent heterogeneity, including cross-sequence (\eg, distinct contrasts for the same tissue in T1/T2 sequences) and cross-site (\eg, contrast of the same tissue in the same sequence varies  with acquisition scanner and setup). Moreover, subject-dependent physiological changes also lead to domain shift. For example, contrasts of white matter and grey matter vary while the human brain undergoes significant growth and expansion within both cortical and subcortical regions during early postnatal years~\cite{gilmore2018imaging}, which contributes to the cross-age domain shift (\hyperref[fig:overview]{Fig.1a}).

The prevalent heterogeneities in medical images lead to suboptimal performance when deep neural networks trained in one source domain are applied to another target domain. To address this challenge, we introduce a \textit{unified} unsupervised domain adaptation (UDA) framework for volumetric and heterogeneous medical image segmentation, named Masked Autoencoding and Pseudo-Labeling Segmentation (\textbf{MAPSeg}). To the best of our knowledge, MAPSeg is the first framework that can be used in \textbf{centralized}, \textbf{federated}, and \textbf{test-time} UDA for volumetric medical image segmentation while maintaining comparable performance. This versatility is particularly advantageous in the field of medical image segmentation, where data sharing is restricted and annotations are expensive. While centralized UDA delivers the best performance in most cases, the strict requirement of co-located data limits its application in multi-institutional studies due to regulations such as the Health Insurance Portability and Accountability Act (HIPAA) and EU General Data Protection Regulation (GDPR)~\cite{MAL-083, insight_from_GDPR}. MAPSeg circumvents this restriction with federated and test-time adaptation, enabling clinical and research collaboration across different medical centers. In contrast, some previous studies, despite showing promising results in one scenario, may become infeasible or suffer significant performance drop in others due to the requirement for co-located data or synchronous adaptation.

In addition, we conduct extensive experiments on a private infant brain MRI dataset, which includes expert-provided annotations, to evaluate MAPSeg on cross-sequence, cross-site, and cross-age adaptation tasks. MAPSeg is also compared with previously reported state-of-the-art (SOTA) results on a public cardiac CT $\rightarrow$ MRI segmentation task. MAPSeg consistently outperforms previous SOTA methods by a large margin (10.5 Dice improvement on the private MRI dataset and 5.7 on the public CT-MRI dataset in the centralized UDA setting). While previous studies have separately explored one of the abovementioned domain shifts~\cite{iseg,9741336, Cui_structure_driven}, they may not generalize to others. For example, cross-age domain shift is mainly composed of changes in brain size and contrast, and methods based on image-translation fail to handle it as they also change the size when translating data from target domain to source domain, leading to segmentation errors. We systematically evaluate MAPSeg across various domain shifts and imaging modalities, demonstrating its consistent and generalizable effectiveness. 

Moreover, in all three UDA settings, MAPSeg does not rely on any target labels for model validation and selection. On the contrary, some previous studies on cardiac CT $\rightarrow$ MRI segmentation~\cite{8988158, Chen_Dou_Chen_Qin_Heng_2019} validate and select the best model using labeled target data, which may not be readily available in real-world problems. We demonstrate that MAPSeg surpasses the previous SOTA results without using any target label for validation, and the performance drop between using and without using target label is minor (0.9 mean Dice). This further justifies its practical value in real-world medical image segmentation tasks. The  contributions of this study are multi-fold: 
\begin{enumerate}
  \item We propose MAPSeg, a unified UDA framework capable of handling various domain shifts in medical image segmentation. 
  \item MAPSeg is suitable for universal UDA scenarios, suggesting its versatility and practical value for real-world problems.  
  \item MAPSeg is extensively evaluated on both private and public datasets, outperforming previous SOTA methods by a large margin. We conduct detailed ablation studies to investigate the impact of each component of MAPSeg.
\end{enumerate}

\section{Related Work}
\label{sec:works}
\subsection{Masked Image Modeling}
Masked image modeling (MIM) represents a category of methods that learn representations from corrupted or incomplete images~\cite{Doersch_2015_ICCV, pmlr-v119-chen20s, pmlr-v119-henaff20a, Pathak_2016_CVPR}, and can naturally serve as a pretext task for self-supervised learning. For example, masked autoencoder (MAE) trains an encoder by reconstructing missing regions from a masked image input and has demonstrated improved generalization and performance in downstream tasks~\cite{He_2022_CVPR, Xie_2022_CVPR, 10.1007/978-3-031-20056-4_18, Xie_2023_CVPRa, Xie_2023_CVPRb,Kong_2023_CVPR, Tang_2022_CVPRb}. MAPSeg heavily relies on MIM, leveraging MAE and masked pseudo-labeling (MPL), to achieve versatile UDA.

\subsection{Pseudo-Labeling}
Pseudo-labeling facilitates learning from limited or imperfect data and is prevalent in semi- and self-supervised learning~\cite{lee2013pseudo,Hu_2021_CVPR, Petrovai_2022_CVPR}. Consistency regularization is widely used in pseudo-label learning~\cite{NIPS2016_30ef30b6, laine2016temporal}, which is a scheme that forces the model to output consistent prediction for inputs with different degrees of perturbation (\eg, weakly- and strongly-augmented images). Mean Teacher~\cite{NIPS2017_68053af2}, a teacher-student framework that generates pseudo-labels from the teacher model (which is a temporal ensembling of the student model), is also a common strategy. In this work, we utilize the teacher model to generate pseudo labels based on complete images and guide the learning of student model on masked images. 

\subsection{Unsupervised Domain Adaptation}
Discrepancy minimization, adversarial learning, and pseudo-labeling are the three main directions explored in UDA. Previous studies have explored minimizing the discrepancy between source and target domains within different spaces, such as input~\cite{Chen_Dou_Chen_Qin_Heng_2019, 9741336, pmlr-v80-hoffman18a}, feature~\cite{he2021autoencoder, 10021602, 8764342, Du_2019_ICCV,NIPS2016_ac627ab1, Feng_Ju_Wang_Song_Zhao_Ge_2023}, and output spaces~\cite{10.1007/978-3-030-58555-6_42, 8954439}, and they sometimes overlap with approaches base on adversarial learning as the supervisory signal to align two distributions may come from statistical distance metrics~\cite{pmlr-v37-long15,NIPS2004_96f2b50b} or a discriminator model~\cite{8764342, 8954439, pmlr-v80-hoffman18a}. Meanwhile, self-training with pseudo-label is also a prevalent technique~\cite{Zhang_2021_CVPR, zheng2021rectifying, Zou_2019_ICCV} and has shown significant improvement on natural image segmentation~\cite{hrda, daformer,Hoyer_2023_CVPR}. Hoyer \etal~\cite{Hoyer_2023_CVPR} proposed masked image consistency as a plug-in to improve previous UDA baselines. In contrast, MAPSeg leverages the synergy between MAE and MPL, and employs MPL as a standalone component for various scenarios.

In this study, we exploit the vanilla pseudo-labeling with three straightforward yet crucial measures to stabilize the training. We hypothesize that random masking is an ideal strong perturbation for consistency regularization in pseudo-labeling, and the model pretrained via MAE can be efficiently adapted to infer semantics of missing regions from visible patches. This hypothesis is justified in Sec. \hyperref[tab:ablation]{4.3}. In addition, we leverage the anatomical distribution prior in medical images and make predictions jointly based on local and global contexts, which also help mitigate the pseudo-label drifts. We demonstrate the superior performance and versatility of MAPSeg in different UDA scenarios in the following sections.

\subsection{Federated Learning}
Federated Learning (FL) is a distributed learning paradigm that aims to train models on decentralized data~\cite{mcmahan2017communication}. FL has attracted great attention in the research community in the last few years and numerous works have focused on the key challenges raised by FL such as data/system heterogeneity~\cite{Tang_2022_CVPR} and communication/computation efficiency~\cite{fedmask}. By virtue of keeping privacy-sensitive medical data local, FL has been adopted for various medical image analysis tasks~\cite{guan2023federated}. Sheller \etal~\cite{sheller} pioneered FL for brain tumor segmentation on multimodal brain scans in a multi-institutional collaboration and showed its promising performance compared to centralized training. Yang \etal~\cite{YANG2021101992} proposed a federated semi-supervised learning framework for COVID-19 detection that relaxed the requirement for all clients to have access to ground truth annotations. FedMix~\cite{FedMix} further alleviated the necessity for all clients to possess dense pixel-level labels, allowing users with weak bounding-box labels or even image-level class labels to collaboratively train a segmentation model. In contrast, MAPSeg assumes all clients have completely unlabeled data when extended to federated UDA scenario. Mushtaq \etal~\cite{DBLP:conf/isbi/MushtaqBDA23} proposed a Federated Alternate Training (\textit{FAT}) scheme that leverages both labeled and unlabeled data silos. It employs mixup~\cite{Mixup} and pseudo-labeling to enable self-supervised learning on the unlabeled participants. MAPSeg, on the other hand, adopts masked pseudo-labeling and global-local feature collaboration for adapting to unlabeled target domains. Yao \etal~\cite{fmtda} introduced the federated multi-target domain adaptation problem and a solution termed \textit{DualAdapt}. It decouples the local-classifier adaptation with client-side self-supervised learning from the feature alignment via server-side mixup and adversarial training. MAPSeg addresses the same federated multi-target UDA problem, and we compare our results to those of \textit{FAT} and \textit{DualAdapt} in Sec. \hyperref[exp:fl]{4.3}.

\subsection{Test-Time UDA}
While federated UDA eases the constraint of centralized data, its learning paradigm still requires synchronous learning across server and clients. Test-time UDA~\cite{chen2021source, NEURIPS2022_bcdec1c2, he2021autoencoder, karani2021test,10.1145/2939672.2939716, Liu_2021_CVPR} assumes the unavailability of source-domain data when adapting to target domains. This assumption significantly limits the applicability of methods based on image translation, adversarial learning, and feature distribution alignment which require simultaneous access to both source and target data. Gandelsman \etal~\cite{NEURIPS2022_bcdec1c2} explored using MAE retraining during test-time to improve classification without employing pseudo labeling. Chen \etal~\cite{chen2021source} proposed using prototype and uncertainty estimation for denoised pseudo labeling of 2D fundus images. Karani \etal~\cite{karani2021test} designed a 2D denoising autoencoder to refine pseudo labels. He \etal~\cite{he2021autoencoder} employed AE during test-time to align source and target feature distributions by minimizing AE reconstruction loss. We demonstrate that, with slight performance drop on source domain, MAPSeg can be extended to test-time UDA with comparable performance to that of centralized UDA on target domain (\cref{sec.ttdaresults}).
\section{Methods}
\label{sec:methods}
\subsection{Preliminary}
\begin{figure*}
\centering
\includegraphics[width=0.85\linewidth]{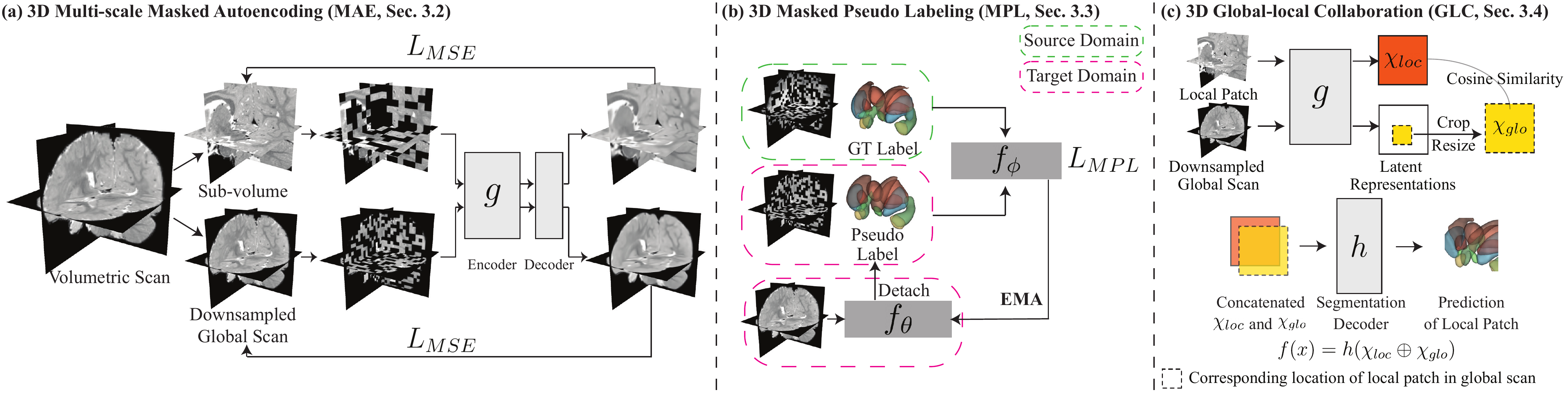}
\caption{Components of the proposed MAPSeg framework. (a) 3D multi-scale masked autoencoding. (b) 3D masked pseudo labeling in source and target domains. (c) 3D Global-local collaboration.} 
\label{fig2}
\end{figure*}

In this section, we introduce each component of MAPSeg (\hyperref[fig2]{Fig.2}) and how MAPSeg can serve as a unified solution to centralized, federated, and test-time UDA (\hyperref[fig:overview]{Fig.1b}). We deploy MAPSeg for domain adaptative 3D segmentation of heterogeneous medical images and it consists of three components: (1) 3D masked multi-scale autoencoding for self-supervised pre-training, (2) 3D masked pseudo-labeling for domain adaptive self-training, and (3) global-local feature collaboration to fuse global and local contexts for the final segmentation task. The hybrid cross-entropy and Dice loss (\hyperref[eq:L_seg]{Eq.1}) is often adopted for regular supervised segmentation training, and we employ it as the basic component of the objective functions for MAPSeg:
\begin{equation}
    \label{eq:L_seg}
    \mathcal{L}_{seg}(\hat{y},y) = -\frac{1}{n}\sum_i\sum_jy_{i,j}\log(\hat{y}_{i,j}) -\frac{2\sum y\hat{y}+\epsilon}{\sum y+\sum \hat{y}+\epsilon}
\end{equation}
where $n$ denotes the number of pixels, $y_{i,j}$ and $\hat{y}_{i,j}$ represent the ground truth label and predicted probability for the $i$th pixel to belong to the $j$th class, and $\epsilon$ is used to prevent zero-division. 

In the following sections, notations are defined as: $x$ and $y$ indicate the original image and label of the randomly sampled local patch; $X$ and $Y$ refer to downsampled global scan and label; the subscripts $s$ and $t$ refer to the source and target domains, respectively; the superscript $M$ indicates the image is masked (\eg, $x_t^M$ refers to a masked local patch from the target domain).

\subsection{3D Multi-Scale Masked Autoencoder (MAE)}
In this study, we propose a 3D variant of MAE using a 3D CNN backbone (\hyperref[fig2]{Fig.2a}). The detailed configuration can be found in Appendix \cref{sec:archite}. Training is jointly performed on two image sources with identical size ($96^3$ voxels): local patches $x$ randomly sampled from the volumetric scan, and the whole scan downsampled to the same size, denoted as $X$. 
Both $x$ and $X$ are masked before feeding into the MAE: $x$ is divided into non-overlapping 3D sub-patches with size $8^3$, of which 70\% are masked out randomly based on a uniform distribution (\hyperref[fig2]{Fig.2a}); The same procedure is applied to $X$ with patch size $4^3$ since it contains a larger field-of-view (FOV). The masked versions of $x$ and $X$ are denoted as $x^M$ and $X^M$, respectively. We train the MAE encoder and decoder to reconstruct $x/X$ based on $x^M/X^M$ using mean squared error on the masked-out regions as the objective function.

\subsection{3D Masked Pseudo-Labeling (MPL)}
MPL uses a teacher-student framework which is a standard strategy in semi-/self-supervised learning~\cite{grill2020bootstrap,NIPS2017_68053af2} to provide stable pseudo labels on an unlabeled target domain during training. 
After MAE pre-training, we keep the MAE encoder $g$ and append a segmentation decoder $h$ to build the segmentation model $f=h\circ g$ (\hyperref[fig2]{Fig.2b-c}). Given an input image $x_s$ and label $y_s$ from the source domain and an input image $x_t$ from the target domain, the teacher model $f_\theta$ takes as input the target image $x_t$ and generates pseudo labels $f_\theta(x_t)$, with gradient detached. The student model $f_\phi$ is then optimized by minimizing the segmentation loss between the predictions of $x_t^M$/$x_s^M$ and $f_{\theta}(x_t)$/$y_s$, which can be formulated as:  
\begin{equation}
\label{eq:L_mpl}
\mathcal{L}_{MPL} = \mathcal{L}_{Seg}(f_{\phi}(x_t^M),f_{\theta}(x_t))+\beta\mathcal{L}_{Seg}(f_{\phi}(x_s^M),y_s)
\end{equation}
where $\beta$ is the weight of source prediction and set as 0.5. 
The teacher model's parameters $\theta$ are then updated during training via exponential moving average (EMA) based on the student model's parameters $\phi$~\cite{NIPS2017_68053af2}.

\begin{equation}
\label{eq:ema_update}
\theta_{t+1} \gets \alpha \theta_{t} + (1-\alpha)\phi_t, 
\end{equation}
where $t$ and $t+1$ indicate training iterations and $\alpha$ is the EMA update weight. For model initialized from the large-scale MAE pretraining, we set $\alpha$ as 0.999 during the first 1,000 steps and 0.9999 afterwards. For model pretrained on small-scale source and target datasets (\eg, only dozens of scans), we set $\alpha$ as 0.99 during the first 1,000 steps, 0.999 during the next 2,000 steps, and 0.9999 for the remaining training. The teacher model $f_{\theta}$ is initialized with student model's parameters $\phi$ after some warm-up training (\eg, 1,000 iterations) on the source-domain data. 

\subsection{3D Global-Local Collaboration (GLC)}
Directly applying MPL for UDA segmentation with large domain shift (\eg, cross-modality/sequence) may lead to unreliable pseudo-label and disrupt the training. Therefore, we design a GLC module (\hyperref[fig2]{Fig.2c}) to improve pseudo-labeling by leveraging the spatial global-local contextual relations induced by the inherent anatomical distribution prior in medical images. With the image encoder pretrained to extract image features at both local and global levels during multi-scale MAE, we take advantage of the global-local contextual relations by concatenating local and global semantic features in the latent space and make prediction based on the fused features. We differ from previous study~\cite{Chen_2019_CVPR} by only applying GLC on the output of the encoder $g$ instead of all layers to save computation cost and employing a different regularization to prevent segmentation decoder from predicting solely based on local features. 

In GLC, a binary mask $M$ is used to indicate the corresponding location of the local patch $x$ inside the downsampled global volume $X$. The encoder $g$ takes as input $x$ and $X$ and generates the local latent feature $\chi_{loc} = g(x)$ as well as cropped and resized global latent feature $\chi_{glo}=\mathit{upsample}(M \odot g(X))$, where $\odot$ indicates cropping $g(X)$ based on $M$ followed by upsampling to match the spatial size of $\chi_{loc}$. Therefore, segmenting a local patch $x$ can be rewritten as $f(x)=h(\chi_{loc}\oplus\chi_{glo})$, where $\oplus$ is the concatenation along channel dimension (\hyperref[fig2]{Fig.2c}). In addition, $f$ is also trained on downsampled global volume $X$ with $\mathcal{L}_{Seg}(f(X),Y)$), in which the global latent feature $g(X)$ is duplicated and $f(X) = h(g(X)\oplus g(X))$, to prevent model from solely relying on local semantic features and encourage the encoder to extract meaningful semantic features from both local and global levels.

We also add a regularization term between the $\chi_{loc}$ and $\chi_{glo}$ to maintain their similarity following~\cite{Chen_2019_CVPR}. Instead of the $\mathcal{L}_2$ regularization used in~\cite{Chen_2019_CVPR}, we maximize the cosine similarity between the $\chi_{loc}$ and $\chi_{glo}$ as:
\begin{equation}
\mathcal{L}_{cos}(x, X) = 1 - \frac{\chi_{loc}\cdot\chi_{glo}}{\max(\| \chi_{loc} \|_2, \| \chi_{glo} \|_2, \epsilon)}
\end{equation}
where $\epsilon$ is used to prevent zero-division. The loss function for GLC calculated on the source data is formulated as: 
\begin{align}
\label{eq.L_gs}
\mathcal{L}_{GLC}^{S} &= \gamma(\mathcal{L}_{Seg}(f_{\phi}(X_s),Y_s)+\mathcal{L}_{Seg}(f_{\phi}(X_s^M),Y_s))
\nonumber\\
&+\delta(\mathcal{L}_{cos}(x_s, X_s) + \mathcal{L}_{cos}(x_s^M, X_s^M))
\end{align}
where $\gamma$ and $\delta$ are the weights of the auxiliary global loss and cosine similarity, and set as $\gamma=0.05$ and $\delta= 0.025$ in our experiments. Similarly, the GLC loss is also calculated on the target data based on pseudo-label $f_{\theta}(X_t)$ and formulated as:
\begin{align}
\label{eq.L_gt}
\mathcal{L}_{GLC}^{T} &= 2\gamma\mathcal{L}_{Seg}(f_{\phi}(X_t^M),f_{\theta}(X_t)) + 2\delta\mathcal{L}_{cos}(x_t^M, X_t^M)
\end{align}
Therefore, the overall loss function of GLC is:
\begin{align}
\label{eq.L_global}
\mathcal{L}_{GLC} &= \mathcal{L}_{GLC}^{S}+\mathcal{L}_{GLC}^{T}
\end{align}
With the regular fully-supervised segmentation loss on source data $\mathcal{L}_{FSS} = \beta\mathcal{L}_{Seg}(f_{\phi}(x_s),y_s)$, where $\beta$ is defined as in \hyperref[eq:L_mpl]{Eq.2}, the overall objective function $\mathcal{L}$ for centralized UDA is formulated as:
\begin{equation}
\label{eq.L_center}
\mathcal{L} = \mathcal{L}_{FSS}+\mathcal{L}_{MPL}+\mathcal{L}_{GLC}
\end{equation}
It is clear that \hyperref[eq.L_center]{Eq.8} requires centralized and synchronous access to source and target data. In the section \hyperref[sec.fuda]{3.5} and \hyperref[sec.ttuda]{3.6}, we demonstrate how MAPSeg can be adapted to federated (decentralized and synchronous access to data) and test-time (decentralized and asynchronous access to data) UDA scenarios. 

\subsection{Extension to Federated UDA}
\label{sec.fuda}
In reality, labeled source-domain data and unlabeled target-domain data are often collected at different sites. We consider a practical scenario where a server (\eg a major hospital) hosts potentially large amount of both labeled and unlabeled scans, and distributed clients (\eg clinics or imaging sites) possess only unlabeled images. This is an under-explored scenario as FL typically assumes either fully or partially labeled data from all clients. We extend MAPSeg to solve this federated multi-target UDA problem according to the details in Algorithm 1 of Appendix \cref{sec:recipe}. Specifically, the server updates the student model $f_\phi$ by minimizing the loss for the labeled source-domain data $D_S$:
\begin{align}
    \mathcal{L}_s 
    &= \beta(\mathcal{L}_{seg}(f_\phi(x_s), y_s)+\mathcal{L}_{seg}(f_\phi(x_s^M), y_s)) \nonumber\\
    &+\gamma(\mathcal{L}_{seg}(f_\phi(X_s), Y_s)+\mathcal{L}_{seg}(f_\phi(X_s^M), Y_s)) \nonumber\\
    &+ \delta(\mathcal{L}_{cos}(x_s, X_s) + \mathcal{L}_{cos}(x_s^M, X_s^M)) \label{eq:loss_server}
\end{align}
The clients update the student model $f_\phi$ by minimizing the loss for its own unlabeled target-domain data $D_T^k$:
\begin{align}
    \mathcal{L}_u
    &= \beta(\mathcal{L}_{seg}(f_\phi(x_t^M), f_\theta(x_t))+\mathcal{L}_{seg}(f_\phi(x_t), f_\theta(x_t))) \nonumber\\
    &+ \gamma(\mathcal{L}_{seg}(f_\phi(X_t^M), f_\theta(X_t))+\mathcal{L}_{seg}(f_\phi(X_t), f_\theta(X_t))) \nonumber\\
    &+ \delta(\mathcal{L}_{cos}(x_t, X_t) + \mathcal{L}_{cos}(x_t^M, X_t^M)) \label{eq:loss_client}
\end{align}
Comparing to the centralized UDA loss (\hyperref[eq.L_center]{Eq.8}), we decompose it into two components: fully supervised loss for server training (\hyperref[eq:loss_server]{Eq.9}) and self-supervised loss for client updates (\hyperref[eq:loss_client]{Eq.10}), which avoids the need for centralized data. After each local update, each client sends the EMA teacher model parameters $\theta$ to the server for aggregation following typical federated averaging\cite{mcmahan2017communication}.

\subsection{Extension to Test-time UDA}
\label{sec.ttuda}
Test-time UDA often involves two separate stages of training, including the source-only training at one center and the target-only finetuning at another site. In the federated UDA setting, \hyperref[eq:loss_server]{Eq.9} and \hyperref[eq:loss_client]{Eq.10} are jointly used to update the server model through synchronous federated averaging after each round. We can further ease the constraint of synchronous communication between source and target sites by training $f_\phi$ on the source data using \hyperref[eq:loss_server]{Eq.9} for some (\eg 1,000) warm-up steps before distributing the model parameters $\phi$ to the target site for initializing the teacher model $f_\theta$. On the target site, $f_\theta$ provides stable pseudo-labels to guide the self-supervised training with \hyperref[eq:loss_client]{Eq.10} and is updated by the EMA of $\phi$ following \hyperref[eq:ema_update]{Eq.3}. We find that in this asynchronous setting MAPSeg still performs well on the target-domain data, albeit with a minor performance tradeoff on the source-domain data (see \hyperref[tab:testtime]{Tab.3}).

%-------------------------------------------------------------------------
\subsection{Implementation Details} \label{section:2.1}

\noindent\textbf{Model architecture and implementation.} We implement the encoder backbone $g$ using 3D-ResNet-like CNN. The segmentation decoder $h$ is adapted from DeepLabV3~\cite{chen2017rethinking}. The framework is implemented using PyTorch. More details of the model and the training procedure are provided in Appendix \cref{sec:archite} and \cref{sec:recipe}. 

\noindent\textbf{Selecting the best model.} For choosing the best model during training, some studies choose to train for fixed iterations and use the last checkpoint. On the other hand, some of the previous UDA studies~\cite{8988158,Chen_Dou_Chen_Qin_Heng_2019} face a dilemma in selecting the best model during training by validating against a hold-out portion of target-domain labels, which is unrealistic as UDA assumes full absence of target labels. We demonstrate that MPL not only provides an efficient pathway to domain adaptative segmentation but also serves as an indicator of how well the model is being adapted to the target domain. We validate the model after each epoch and the best model is selected based on the score: 
$\mathit{Score}=\mathit{Dice}_{Src}-0.5\times\overline{\mathcal{L}_{Seg}}(f_{\phi}(x_t^M),f_{\theta}(x_t))$, where $\mathit{Dice}_{Src}$ is the Dice score on source-domain validation set and $\overline{\mathcal{L}_{Seg}}(f_{\phi}(x_t^M),f_{\theta}(x_t))$ is the mean of $\mathcal{L}_{Seg}(f_{\phi}(x_t^M),f_{\theta}(x_t))$ during the last training epoch. From \hyperref[eq:L_seg]{Eq.1}, it is clear that $ \lim_{\hat{y}\to y} \mathcal{L}_{seg}(\hat{y},y)=-1$, therefore, $Score$ has an upper bound of $1.5$. We demonstrate in \hyperref[tab:cardiac]{Tab.4} that the difference between validation using target labels versus $Score$ is acceptable (81.2 vs. 80.3). Even without accessing target labels for validation, MAPSeg still surpasses the previous SOTA results that use target labels for validation. It is worth noting that we only use target labels for validation in \hyperref[tab:cardiac]{Tab.4} for a fair comparison with previously reported results; other results presented use $Score$ for validation by default. For federated and test-time UDA, $\mathit{Score} = -\overline{\mathcal{L}_{Seg}}(f_{\phi}(x_t^M),f_{\theta}(x_t))$.

\section{Experiments and Results}
\subsection{Datasets}
\label{sec:braindata}
\noindent\textbf{Brain MRI Datasets.} We include 2,421 (1,163 T1w) brain MRI scans acquired from newborn to toddler in this study. Among them, 2,306 are unannotated scans dedicated for the 3D multi-scale MAE pretraining. These MRI scans are acquired from multiple sites with different sequence parametrization and scanner types. All scans are preprocessed with skull stripping~\cite{hoopes2022synthstrip} and bias-field correction~\cite{tustison2010n4itk}. These MRI brain scans were acquired worldwide, and detailed descriptions can be found in Appendix \cref{sec:datadetail}. 

To evaluate cross-sequence/site/age UDA segmentation for seven subcortical regions (\ie, hippocampus (HC), amygdala (AD), caudate (CD), putamen (PT), pallidum (PD), thalamus (TM), and accumbens (AB)), our analysis include manual segmentation of 115 scans. They comprise independent subjects from the BCP cohort (\textit{BCP50}) with private expert segmentation for both T1w and T2w scans (acquired from 0 to 24 months postnatal age); 5 newborn scans from the ECHO cohort (\textit{ECHO5}) with private expert segmentation; and 10 newborn scans from the M-CRIB project (\textit{MCRIB10}) with publicly available segmentation~\cite{mcrib}.

\noindent\textbf{Cardiac CT-MRI Dataset.}
Following the previous studies~\cite{8988158, Chen_Dou_Chen_Qin_Heng_2019}, we include 40 independent scans (20 CT and 20 MRI) of cardiac regions from Multi-Modality Whole Heart Segmentation (MMWHS) Challenge 2017 dataset~\cite{ZHUANG201677, 9965747, 8458220} with ground truth labels of ascending aorta (AA), left atrium blood cavity (LAC), left ventricle blood cavity (LVC), and myocardium of the left ventricle (MYO). Similarly, we apply bias-field correction to the MRI scans.

%-------------------------------------------------------------------------
\begin{table}
    \caption{Performance of centralized UDA on brain MRI segmentation.}
    \vspace{-1em}
    \centering
    \begin{center}
        \resizebox{0.88\linewidth}{!}{%
            \begin{tabular}{c|cccccccc}
                \toprule
                \multicolumn{9}{c}{\textbf{Cross-Sequence}}\\	
                \hline
                \multirow{2}{*}{Method} &\multicolumn{8}{c}{Dice(\%) $\uparrow$} \\
                \cline{2-9}
                &HC &AD &CD &PT &PD&TM &AB &Avg \\
                
                \hline
                AdvEnt\cite{8954439} &56.7&52.7&66.7&66.1&61.8&74.1&40.1&59.8\\
                DAFormer\cite{daformer}&40.5&53.3&62.2&64.7&45.9&61.8&39.9&52.6\\
                HRDA\cite{hrda}&42.6&37.7&66.5&71.9&0.0&67.6&0.3&40.9\\
                MIC\cite{Hoyer_2023_CVPR}&40.3&47.0&72.5&52.9&0.0&62.1&0.0&39.3\\
                DAR-UNet\cite{9741336} &61.3&65.2&76.7&75.8&68.1&82.0&48.4&68.2\\
                \textbf{MAPSeg (Ours)} &\textbf{70.3}&\textbf{73.2}&\textbf{81.4}&\textbf{83.9}&\textbf{76.5}&\textbf{89.6}&\textbf{69.2}&\textbf{77.7}\\
                \bottomrule

                \multicolumn{9}{c}{\textbf{Cross-Site}}\\	
                \hline
                \multirow{2}{*}{Method} &\multicolumn{8}{c}{Dice(\%) $\uparrow$} \\
                \cline{2-9}
                &HC &AD &CD &PT &PD&TM &AB &Avg \\
                
                \hline
                AdvEnt\cite{8954439} &27.1&6.7&21.0&23.1&12.5&36.0&20.5&21.0\\
                DAFormer\cite{daformer}&40.0&45.8&75.3&70.0&68.4&64.0&51.3&59.3\\
                HRDA\cite{hrda}&30.9&44.3&80.8&79.8&66.4&83.0&53.4&62.7\\
                MIC\cite{Hoyer_2023_CVPR}&48.1&36.2&67.7&82.8&\textbf{69.5}&66.8&52.3&60.5\\
                DAR-UNet\cite{9741336} &51.9&43.6&69.8&55.2&55.5&81.2&45.8&57.6\\
                \textbf{MAPSeg (Ours)} &\textbf{70.0}&\textbf{53.5}&\textbf{85.6}&\textbf{85.4}&67.9&\textbf{88.1}&\textbf{61.4}&\textbf{73.1}\\
                \bottomrule
                \multicolumn{9}{c}{\textbf{Cross-Age}}\\	
                \hline
                \multirow{2}{*}{Method} &\multicolumn{8}{c}{Dice(\%) $\uparrow$} \\
                \cline{2-9}
                &HC &AD &CD &PT &PD&TM &AB &Avg \\
                
                \hline
                AdvEnt\cite{8954439} &58.7&54.1&44.0&63.8&56.9&78.0&30.9&55.2\\
                DAFormer\cite{daformer}&30.2&65.7&72.7&55.8&38.4&88.8&57.3&58.4\\
                HRDA\cite{hrda}&48.6&66.6&81.9&67.7&35.7&74.1&56.0&61.5\\
                MIC\cite{Hoyer_2023_CVPR}&61.3&66.0&80.9&\textbf{73.4}&44.3&76.1&51.0&64.7\\
                DAR-UNet\cite{9741336} &58.8&56.3&64.4&64.5&53.6&82.6&28.6&58.8\\
                \textbf{MAPSeg (Ours)} &\textbf{75.8}&\textbf{76.7}&\textbf{83.1}&71.4&\textbf{58.2}&\textbf{90.7}&\textbf{70.1}&\textbf{75.2}\\
                \bottomrule
            \end{tabular}}
    \end{center}
    \label{tab:centralizedUDA}
    \vspace{-1em}
\end{table}

\subsection{Dataset Partition}

\noindent\textbf{Pretraining.} For multi-scale MAE pretraining on brain MRI scans, we have four models pretrained on different amounts of data to investigate the influence of pretraining data size. The model pretrained on large-scale data takes advantage of all 2,306 unannotated scans introduced in \cref{sec:braindata}. Since there is no overlapping with the annotated scans, the pretrained model can be directly applied to all downstream UDA tasks (\ie, cross-site/age/sequence). We also pretrain the model solely relying on source and target training data of each task.

For multi-scale MAE pretraining on cardiac CT-MRI scans, the model is only pretrained on training scans of source (16 CT scans) and target (16 MRI scans) domains, following the partition adopted by previous studies.

\noindent\textbf{Cross-Sequence UDA segmentation of brain.}
The model is trained on T1w MRI scans (source domain) and tested on T2w MRI scans (target domain). 
The \textit{BCP50} dataset is randomly split into two non-overlapping subsets of 25 subjects per each. The model is trained on T1w scans of the first group (source domain 18 scans for training and 7 for validation) and T2w scans of the second group (target domain 15 for training and 10 for testing). The best validation model is then applied to the T2w testing scans. 

\noindent\textbf{Cross-Site UDA segmentation of brain.}
The model is trained on a single site (\textit{BCP50}, source domain) and tested on two other sites (\textit{MCRIB10} and \textit{ECHO5}, target domains). Utilizing 50 T2w MRI scans from BCP as the source domain, we randomly select 40 scans for training and 10 for validation. Six scans from \textit{MCRIB10} and three scans from \textit{ECHO5} are used for UDA training, and remaining scans are used for testing. 

\noindent\textbf{Cross-Age UDA segmentation of brain.}
We also conduct experiments in cross-age segmentation using longitudinal scans from \textit{BCP50}. We set the 24 T2w MRI scans of 12-24 month-old infant as the source domain and 14 T2w MRI scans of 0-6 month-old infants as the target domain. For the source domain, 19 scans are randomly sampled for training and remaining 5 scans are used for validation. For the target domain, 8 scans are used for UDA training and 6 scans are used for testing. 

\noindent\textbf{Cross-Modality UDA segmentation of cardiac.}
For the cardiac scans, for a fair comparison, we follow the same partition employed by the previous studies. We set CT as the source domain and MRI as the target domain, and use 16 CT scans and 16 MRI scans for training, 4 CT scans for validation, and the remaining 4 MRI scans for testing.  

\subsection{Results}

\noindent\textbf{Centralized Domain Adaptation.}
\label{sec.CUDA}
To assess MAPSeg's performance in different UDA tasks for infant brain MRI segmentation, we compare it with methods utilizing adversarial entropy minimization~\cite{8954439}, image translation~\cite{9741336}, and pseudo-labeling~\cite{daformer,hrda, Hoyer_2023_CVPR}. The results are reported in \hyperref[tab:centralizedUDA]{Tab.1}. MAPSeg consistently outperforms its counterparts across all tasks. DAR-UNet ranks second in the cross-sequence task but shows degraded performance in others, partially due to translation error (details in Appendix). Among pseudo-labeling approaches, HRDA and MIC achieve the second best performance in cross-site and cross-age tasks, respectively. However, they fail to segment pallidum and accumbens in the cross-sequence task. A major challenge here is the small size of subcortical regions (accounting for approximately 2\% of overall voxels) and significant class imbalance (\eg, thalamus comprises about 0.8\% of overall voxels, while accumbens accounts for only 0.03\%). This imbalance poses a significant challenge for previous pseudo-labeling methods. Additional visualizations and discussions are available in Appendix \cref{sec:appendixvis}.

\begin{table}
    \caption{Performance of federated UDA on brain MRI segmentation.}
    \vspace{-1em}
    \centering
    \begin{center}
        \resizebox{0.8\linewidth}{!}{%
            \begin{tabular}{c|ccc}
                \toprule
                \multirow{2}{*}{Method} & \multicolumn{3}{c}{Dice(\%) $\uparrow$} \\
                \cline{2-4} & Cross-Sequence & Cross-Site & Cross-Age \\
                \hline
                FAT\cite{DBLP:conf/isbi/MushtaqBDA23} & 27.6 & 63.8 & 69.0 \\
                DualAdapt\cite{fmtda} & 28.4 & 66.1 & 54.8 \\
                \hline
                \textbf{Fed-MAPSeg (ours)} &\textbf{69.9} &\textbf{73.6} &\textbf{71.0} \\
                \bottomrule
            \end{tabular}}
    \end{center}
    \label{exp:fl}
    \vspace{-1em}
\end{table}

\noindent\textbf{Federated Domain Adaptation.}
\label{sec:results_FLUDA}
To evaluate our framework in the federated domain adaptation setting, we designate the labeled source-domain dataset as the server dataset and the unlabeled target-domain datasets as the client datasets. In the cross-sequence setting, the 25 T1w scans of the first group are considered as the server dataset, and the 25 T2w scans of the second group are split roughly equally into three disjoint client datasets. In the cross-site setting, the \textit{BCP50} is considered as the server dataset, and the \textit{ECHO5} and \textit{MCRIB10} naturally serve as two different client datasets. In the cross-age setting, we treat the scans from the first age group as the server dataset, and split the scans from the second age group equally into two client datasets.

\begin{table}
    \caption{Comparison between centralized and test-time UDA on brain MRI segmentation. Performance of source domain are reported on source validation set.}
    \vspace{-1em}
    \centering
    \begin{center}
        \resizebox{0.86\linewidth}{!}{%
            \begin{tabular}{c|cccc|cc}
                \hline
                \multirow{2}{*}{Task} &\multicolumn{2}{c|}{Centralized UDA}&\multicolumn{2}{c|}{Test-time UDA}&\multirow{2}{*}{$\Delta_{Source}$}&\multirow{2}{*}{$\Delta_{Target}$}\\
                \cline{2-5}
                
                &\multicolumn{1}{c}{Source}&\multicolumn{1}{c|}{Target}&\multicolumn{1}{c}{Source} &\multicolumn{1}{c|}{Target} \\
                
                \hline
                X-seq &84.0&77.7&79.2&75.9&-4.8&-1.8\\
                X-age & 85.8 &75.2& 84.2&72.9&-1.6&-2.3\\
                X-site &85.7&73.1& 79.9& 70.3&-5.8&-2.8\\
                \bottomrule

        \end{tabular}}
    \end{center}
    \label{tab:testtime}
    \vspace{-1em}
\end{table}

We compare our Fed-MAPSeg with two other related work, \textit{FAT}~\cite{DBLP:conf/isbi/MushtaqBDA23} and \textit{DualAdapt}~\cite{fmtda}. To our best knowledge, there is no direct comparison from the literature that addresses this challenging federated multi-target unsupervised domain adaptation for 3D medical image segmentation. \textit{FAT}~\cite{DBLP:conf/isbi/MushtaqBDA23} proposes an alternating training scheme between the labeled and unlabeled data silos and adopts a mixup approach to augment the unlabeled input data for self-supervised learning with pseudo-labels. \textit{DualAdapt}~\cite{fmtda} considers a similar single-source to multi-target unsupervised domain adaptation setting, except that it only reports segmentation performance for 2D image datasets such as the DomainNet~\cite{peng2019moment} and CrossCity~\cite{cross_city}. Implementation details for our Fed-MAPSeg as well as the baselines are included in Appendix \cref{sec:baseline}. We report our results in \hyperref[exp:fl]{Tab.2}. Fed-MAPSeg not only outperforms the two baselines by a large margin (esp. in the the cross-sequence setting), it also maintains a fairly close performance compared to the centralized UDA.

\label{sec.ttdaresults} \noindent\textbf{Test-Time Domain Adaptation.} We further extend MAPSeg to Test-time UDA, and the results for different tasks are reported in \hyperref[tab:testtime]{Tab.3}. With decentralized data and asynchronous training, MAPSeg still performs very well in all tasks, with performance drop smaller than 3\% in the target domain. However, we observe a slightly more performance degradation in the source domain (\hyperref[tab:testtime]{Tab.3}), particularly in cross-sequence and cross-site tasks, suggesting that the model suffers from forgetting of the source domain knowledge during test-time UDA. 

\noindent\textbf{Cross-Modality Segmentation of Cardiac.} 
To evaluate the generalizability of MAPSeg, we further conduct experiment for cross-modality cardiac segmentation and the results are reported in \hyperref[tab:cardiac]{Tab.4}. MAPSeg surpasses all previously reported results. Results of MRI $\rightarrow$ CT segmentation can be found in Appendix \cref{sec:mri2ctresult}.

\begin{table}
    \caption{Performance of centralized UDA on cardiact CT$\rightarrow$ MRI segmentation. Underline indicates the target labels are not used for validation.}
    \vspace{-1em}
    \centering
    \begin{center}
        \resizebox{0.78\linewidth}{!}{%
            \begin{tabular}{c|cccccccc}
                \toprule
                \multicolumn{6}{c}{Cardiac CT $\rightarrow$ MRI segmentation}\\	
                \hline
                \multirow{2}{*}{Method} &\multicolumn{5}{c}{Dice(\%) $\uparrow$} \\
                \cline{2-6}
                &AA &LAC &LVC &MYO &Avg \\
                \hline
                PnP-AdaNet\cite{8764342} &43.7&47.0&77.7&48.6&54.3 \\
                SIFA-V1\cite{Chen_Dou_Chen_Qin_Heng_2019} &67.0&60.7&75.1&45.8&62.1 \\
                SIFA-V2\cite{8988158} &65.3&62.3&78.9&47.3&63.4 \\
                DAFormer\cite{daformer} &75.2&59.4&72.0&57.1&65.9 \\
                MPSCL\cite{9672690} & 62.8&76.1&80.5&55.1&68.6\\
                MA-UDA\cite{10273225} &71.0&67.4&77.5&57.1&68.7 \\
                SE-ASA\cite{Feng_Ju_Wang_Song_Zhao_Ge_2023} &68.3&74.6&81.0&55.9&69.9 \\
                FSUDA-V1\cite{Liu_Yin_Qu_Wang_2023} &62.4&72.1&81.2&66.5&70.6\\
                PUFT\cite{10021602} &69.3&77.4&83.0&63.6&73.3\\
                SDUDA\cite{Cui_structure_driven}&72.8&79.3&82.3&64.7&74.8\\
                FSUDA-V2\cite{10261458} &72.5&78.6&82.6&68.4&75.5\\
                
                \hline
                \multirow{2}{*}{\textbf{MAPSeg (Ours)}}
                &\underline{\textbf{78.5}}&\underline{\textbf{81.8}}&\underline{92.1}&\underline{68.8}&\underline{80.3}\\
                 &78.2&\textbf{81.8}&\textbf{92.9}&\textbf{72.0}&\textbf{81.2}\\
                \bottomrule

        \end{tabular}}
    \end{center}
    \label{tab:cardiac}
    \vspace{-1em}
\end{table}
\begin{table}
    \caption{Ablation studies of MAPSeg components on cross-sequence brain MRI segmentation.}
    \vspace{-1em}
    \centering
    \begin{center}
        \resizebox{0.5\linewidth}{!}{%
            \begin{tabular}{cccccc|c}
                \toprule
                \multicolumn{6}{c|}{Components}&\multicolumn{1}{c}{Performance}\\
                    \hline
                \cline{1-7}
                \multicolumn{2}{c|}{MAE} &\multicolumn{2}{c|}{GLC} &\multicolumn{2}{c|}{MPL} &Dice(\%) $\uparrow$\\
                
                \hline
    \multicolumn{2}{c|}{} &\multicolumn{2}{c|}{} &\multicolumn{2}{c|}{} &31.6  \\
     \multicolumn{2}{c|}{\checkmark} &\multicolumn{2}{c|}{} 
     &\multicolumn{2}{c|}{} &51.3 \\
     \multicolumn{2}{c|}{} &\multicolumn{2}{c|}{\checkmark} &\multicolumn{2}{c|}{} &53.0  \\
     \multicolumn{2}{c|}{} &\multicolumn{2}{c|}{} &\multicolumn{2}{c|}{\checkmark} &39.5  \\
     \multicolumn{2}{c|}{} &\multicolumn{2}{c|}{\checkmark} &\multicolumn{2}{c|}{\checkmark} & 59.0  \\
     \multicolumn{2}{c|}{\checkmark} &\multicolumn{2}{c|}{\checkmark} &\multicolumn{2}{c|}{} &71.3  \\
     \multicolumn{2}{c|}{\checkmark} &\multicolumn{2}{c|}{} &\multicolumn{2}{c|}{\checkmark} &75.3  \\
     
     \multicolumn{2}{c|}{\checkmark} &\multicolumn{2}{c|}{\checkmark} &\multicolumn{2}{c|}{\checkmark} &77.7  \\
					
                \bottomrule
        \end{tabular}}
    \end{center}
    \label{tab:ablation}
    \vspace{-1em}
\end{table}
\noindent\textbf{Ablation Studies.}
\label{sec:ablation}
To further investigate each component of MAPSeg, we conduct ablation studies focusing on MAE, GLC, MPL, masking ratio, masking patch size of local patch, and pretraining data size in the context of cross-sequence segmentation. From \hyperref[tab:ablation]{Tab.5}, it is clear that directly applying MPL only brings a minor improvement, suggesting using MPL alone suffers from pseudo-label drifts. By incorporating GLC to leverage global-local contexts, MPL yields better results. MAE pretraining significantly boosts the performance from using MPL alone (39.5 to 75.3), justifying MAE and MPL are complementary parts in MAPSeg. Combining MAE, MPL, and GLC together yields the optimal performance. 

\begin{figure}[t!]
  \centering
     \includegraphics[width=0.9\linewidth]{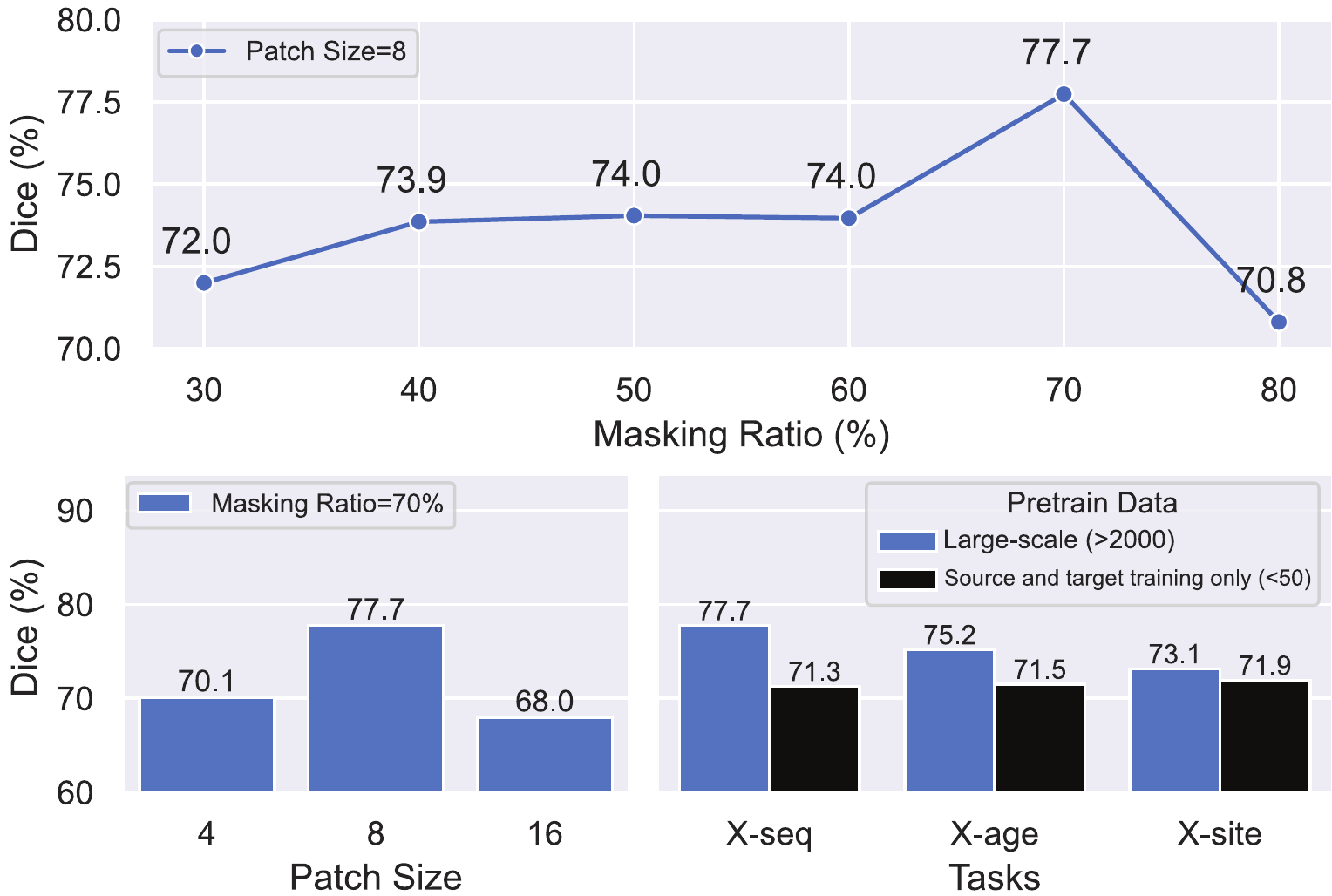}
  \caption{Ablation studies on masking ratio, patch size, and pretrain data. Experiments on masking ratio and patch size are conducted on cross-sequence task.}
  \label{fig:ablation}
  \vspace{-1em}
\end{figure}

The impact of masking ratio and local patch size is reported in \hyperref[fig:ablation]{Fig.3}. The masking ratio and patch size remain the same in MAE and MPL. The results indicate that MAPSeg is more sensitive to patch size. A patch size of 4 or 16 decreases the performance significantly. For the masking ratio, MAPSeg achieves optimal performance when 70\% of the regions are masked out. Additionally, we evaluate model's performance using only source and target training data ($<$ 50 scans) for MAE pretraining, much fewer than the large-scale pretraining ($>$ 2,000 scans). This suggests that, even with dozens of scans involved in MAE, MAPSeg still delivers comparable performance. Another benefit of large-scale pretraining is its immediate applicability to new target domains; the pretrained encoders can be directly employed for MPL, bypassing the need for training from scratch. Additional analyses about sensitivity to other hyperparameters can be found in Appendix \cref{sec:additionres}.

\section{Conclusions}
In this paper, we introduce the MAPSeg framework as a unified UDA framework that works on centralized, federated, and test-time UDA scenarios. We evaluate it under multiple domain shift and adaptation settings, and it outperforms all the baselines in all scenarios.
We conduct extensive ablation study to demonstrate the effectiveness of each component.

\section{Acknowledgements}
This work was supported by NIH grants R00HD103912 (Y.W.), R01HL121270 (R.G.B. \& A.F.L.), R01MH121070 (J.P. \& A.P.J.), and NSF grant CNS-2112562 (H.L.), as well as by Duke Science and Technology (Y.W. \& H.L.).

\newpage

{
    \small
    \bibliographystyle{ieeenat_fullname}
    \bibliography{main}
}
\renewcommand{\figurename}{Suppl.Fig.}
\renewcommand{\tablename}{Suppl.Tab.}
\setcounter{table}{0}
\setcounter{figure}{0}
\setcounter{section}{0}

\clearpage
\setcounter{page}{1}
\maketitlesupplementary

\section{Appendix}
\label{sec:appendix}

\newlength\savewidth\newcommand\shline{\noalign{\global\savewidth\arrayrulewidth
  \global\arrayrulewidth 1pt}\hline\noalign{\global\arrayrulewidth\savewidth}}
\newcommand{\tablestyle}[2]{\setlength{\tabcolsep}{#1}\renewcommand{\arraystretch}{#2}\centering\footnotesize}

\subsection{Model Architecture}
\label{sec:archite}
\newcommand{\blocka}[2]{\multirow{3}{*}{\(\left[\begin{array}{c}\text{3$\times$3, #1}\\[-.1em] \text{3$\times$3, #1} \end{array}\right]\)$\times$#2}
}
\newcommand{\block}[3]{\multirow{3}{*}{\(\left[\begin{array}{c}\text{4$\times$4$\times$4, #1}\\[-.1em] \text{3$\times$3$\times$3, #2}\\[-.1em] \text{3$\times$3$\times$3, #2}\end{array}\right]\)$\times$#3}
}
\newcommand{\blockb}[3]{\multirow{3}{*}{\(\left[\begin{array}{c}\text{3$\times$3$\times$3, #2}\\[-.1em] \text{3$\times$3$\times$3, #2}\\[-.1em] \text{3$\times$3$\times$3, #1}\end{array}\right]\)$\times$#3}
}
\newcommand{\seghead}[1]{\multirow{2}{*}{\(\left[\begin{array}{c}\text{3$\times$3$\times$3, 64}\\[-.1em] \text{1$\times$1$\times$1, cls\_num}\\[-.1em] \end{array}\right]\)$\times$#1}
}
\begin{figure}[b]
  \centering
     \includegraphics[width=1\linewidth]{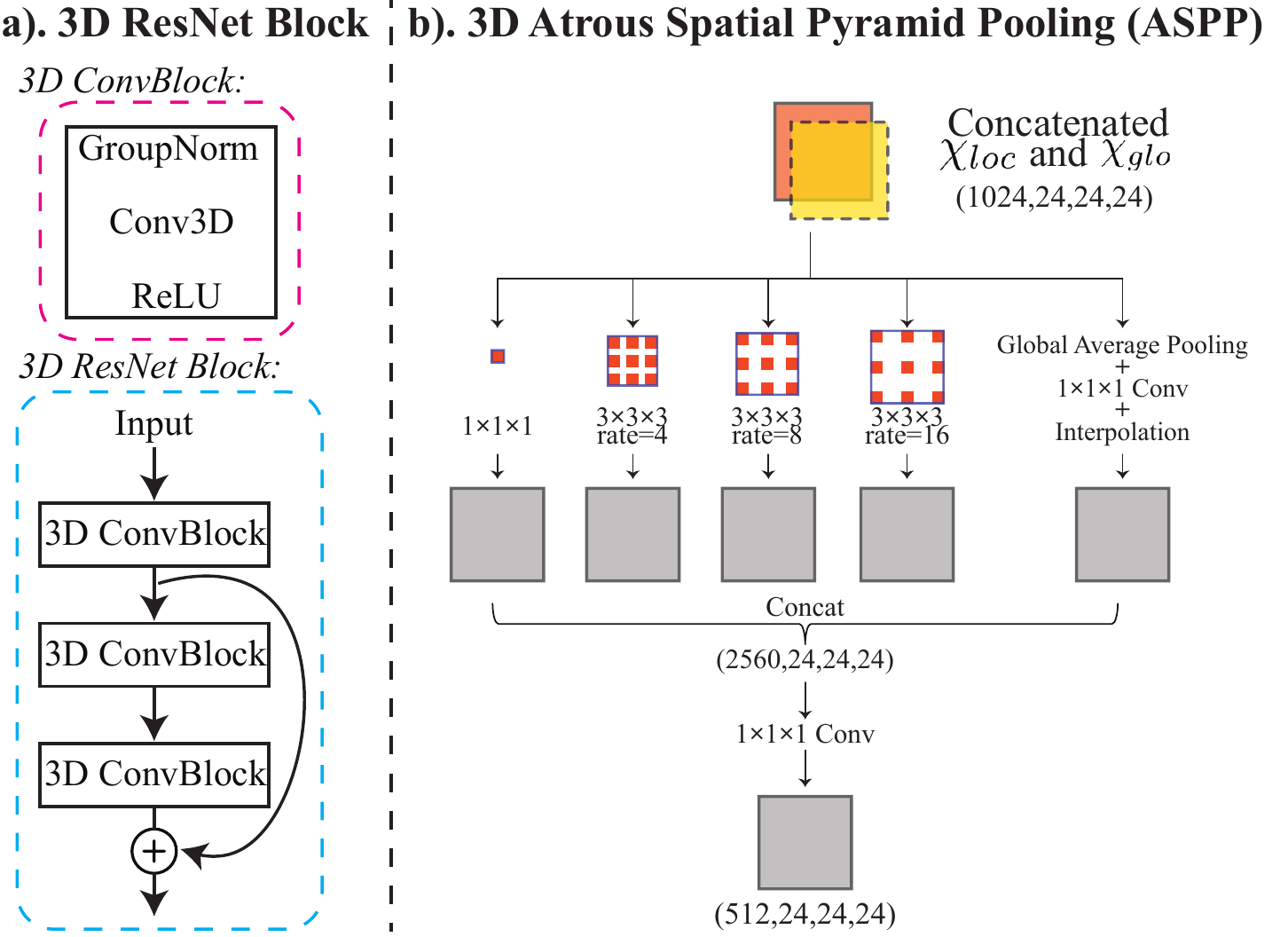}
  \caption{Illustrations of 3D ResNet Block and 3D Atrous Spatial Pyramid Pooling (ASPP) layer.}
  \label{fig:arch}
\end{figure}

MAPSeg is implemented using PyTorch. Detailed configurations of model and training can be found below. 

\noindent\textbf{ 3D Multi-Scale Masked Autoencoder (MAE).} We implement the 3D MAE using 3D ResNet Blocks \cite{he2016deep,wolny2020accurate} instead of Vision Transformers, different from the previous study \cite{He_2022_CVPR}, due to the constraint of GPU memory. The encoder consists of eight 3D ResNet Blocks. The 3D ResNet Block is depicted in \hyperref[fig:arch]{Suppl.Fig.1a}. Following the previous study \cite{He_2022_CVPR}, we adopt an asymmetric design by employing a lightweight decoder (\hyperref[tab:arch]{Suppl.Tab.1}). 

\renewcommand\arraystretch{1.1}
\setlength{\tabcolsep}{3pt}
\begin{table}[t]
\label{tab:arch}
\begin{center}
\resizebox{1\linewidth}{!}{
\begin{tabular}{c|c|c|c}
\hline
\toprule
\multicolumn{4}{c}{\textbf{Encoder}}\\	
\hline
Layer Name & Input Size & Output Size & Architecture \\
\hline
\multirow{3}{*}{enc\_res1} & \multirow{3}{*}{(1,96,96,96)} & \multirow{3}{*}{(512,24,24,24)} &  \block{512}{512}{1} \\
&  &  &  \\
&  &  &  \\
\hline
\multirow{3}{*}{enc\_res2.x} & \multirow{3}{*}{(512,24,24,24)} & \multirow{3}{*}{(512,24,24,24)} &  \blockb{512}{512}{7} \\
&  &  &  \\
&  &  &  \\
\toprule
\multicolumn{4}{c}{\textbf{MAE Decoder}}\\	
\hline
Layer name & Input size & Output size & Architecture \\
\hline
trans\_conv1 & (512,24,24,24) & (32,96,96,96) &  4$\times$4$\times$4, 32, stride 4\\

\hline
\multirow{3}{*}{dec\_res1} & \multirow{3}{*}{(32,96,96,96)} & \multirow{3}{*}{(16,96,96,96)} &  \blockb{16}{16}{1} \\
&  &  &  \\
&  &  &  \\
\hline
final\_recon & (16,96,96,96) & (1,96,96,96) &  3$\times$3$\times$3, 1, stride 1\\
\toprule
\multicolumn{4}{c}{\textbf{Segmentation Decoder}}\\	
\hline
Layer name & Input size & Output size & Architecture \\
\hline
ASPP & (1024,24,24,24) & (512,24,24,24) &  \hyperref[fig:arch]{Suppl. Fig.1b}\\
\hline

trans\_conv2 & (512,24,24,24) & (64,96,96,96) &  4$\times$4$\times$4, 64, stride 4\\
\hline
\multirow{2}{*}{seg\_head} & \multirow{2}{*}{(64,96,96,96)} & \multirow{2}{*}{(cls\_num,96,96,96)} &  \seghead{1} \\
&  &  &  \\
\toprule
\end{tabular}
}
\end{center}
\vspace{-.5em}
\caption{Architectures of different components of MAPSeg. Building blocks ([kernal size, output channels]) are shown in brackets, with the number of blocks stacked. Downsampling is performed by the first block of enc\_res1 with a stride of 4.
}
\vspace{-.5em}
\end{table}

\noindent\textbf{3D Global-Local Collaboration (GLC).} The segmentation backbone (\hyperref[tab:arch]{Suppl.Tab.1}) consists of the pretrained encoder and a segmentation decoder that is adapted from DeepLabV3 \cite{chen2017rethinking}. In the decoding path, we take advantage of the Atrous Spatial Pyramid Pooling (ASPP), which employs dilated convolution at multiple scales and provides access to larger FOV (\hyperref[fig:arch]{Suppl.Fig.1b}). After feature extraction, the GLC module fuses the local and global features and forms a latent representation with a dimension of 1024, which is then fed into the ASPP layer. During training, each local sub-volume with size of $96\times96\times96$ is randomly sampled from global scan. During inference, the final output is formed by sliding window with stride of 80 across entire volumetric scan. 

\subsection{Training Recipe}
\label{sec:recipe}
\textbf{MAE Pretraining.}
For the MAE Pretraining, we follow the training configurations listed in \hyperref[tab:mae_pretrain]{Suppl.Tab.2}. Each mini-batch contains a pair of randomly sampled local patch $x$ and downsampled global scan $X$. The masking patch in \hyperref[tab:mae_pretrain]{Suppl.Tab.2} only applies to $x$ and is always half-sized for $X$ because of the larger FOV. For example, in the ablation study of masking patch size, a masking patch of 16 to $x$ indicates a masking patch of 8 to $X$. We implement the augmentation using TorchIO \cite{PEREZGARCIA2021106236}. During the MAE stage, we employ random 3D affine transformation, with isotropic scaling 75-150\% and rotation [-40\textdegree, 40\textdegree].

\begin{table}[]
\tablestyle{6pt}{1.02}
\scriptsize
\begin{tabular}{c|c}
config & value \\
\shline
masking patch & 8$\times$8$\times$8\\
masking ratio & 70\%\\
optimizer & AdamW \cite{loshchilov2018decoupled}\\
learning rate & $2e^{-4}$\\
weight decay &0.05\\
optim. momentum & $\beta_1, \beta_2=0.9,0.95$\\
\multirow{2}{*}{lr scheduler} & cosine annealing \cite{loshchilov2017sgdr}\\
& $T_{max}$=20, min\_lr=$1e^{-6}$\\
total epochs & 300\\
annealing epochs & last 100\\
batch size & 4\\
iters/epoch & 500\\
aug. prob. & 0.35\\
\multirow{1}{*}{augmentation} & random affine \\
\end{tabular}
\vspace{-.5em}
\caption{MAE Pretraining Configurations}
\label{tab:mae_pretrain} \vspace{-.5em}
\end{table}

\noindent\textbf{Centralized UDA.}
For the centralized UDA on brain MRI segmentation tasks, detailed training configuration can be found in \hyperref[tab:cent_mpl]{Suppl.Tab.3}. Similarly, each mini-batch contains a pair of $x$ and $X$ from the source domain and another pair from the target domain (four 96$\times$96$\times$96 patches). During warmup epochs, the model is only trained on source domain. We utilize $\mathit{Score}$ to select the best model and the patience is set as 50 epochs. For the target domain, we design a similar random 3D affine transformation, with scaling 70-130\% and rotation [-30\textdegree, 30\textdegree]. A stronger augmentation strategy is applied to the source domain, consisting of random affine (scaling 70-140\% and rotation [-30\textdegree, 30\textdegree]), random bias field \cite{811270,SUDRE201750}, and random gamma transformation ($\gamma \in [e^{-0.4}, e^{0.4}]$). For the centralized UDA on public cardiac CT$\rightarrow$MRI segmentation, we use the same configuration except for training epochs of 150 and warmup epochs of 50. For MRI$\rightarrow$CT cardiac segmentation, we use a less aggressive augmentation strategy because MRI is noisier than CT. We set the scaling ratio to 85-115\% and rotation to [-15\textdegree, 15\textdegree] for both source and target domains, and exclude random bias field and gamma transformation. The warmup epoch is set as 70. 

\begin{table}[]
\tablestyle{6pt}{1.02}
\scriptsize
\begin{tabular}{c|c}
config & value \\
\shline
masking patch & 8$\times$8$\times$8\\
masking ratio & 70\%\\
optimizer & AdamW \cite{loshchilov2018decoupled}\\
learning rate & $1e^{-4}$\\
weight decay &0.01\\
optim. momentum & $\beta_1, \beta_2=0.9,0.999$\\
\multirow{2}{*}{lr scheduler} & cosine annealing warm restart \cite{loshchilov2017sgdr}\\
& $T_{0}$=10, $T_{mult}$=2, min\_lr=$1e^{-8}$\\
total epochs & 100\\
warmup epochs & first 10 \\
annealing epochs & all\\
early stop & 50\\
batch size & 1\\
iters/epoch & 100\\
aug. prob. & 0.35\\
\multirow{3}{*}{source aug.} & random affine \\
& random bias field \\
& random gamma trans. \\
\multirow{1}{*}{target aug.} & random affine \\

\end{tabular}
\vspace{-.5em}
\caption{Centralized UDA configurations for brain MRI segmentation.}
\label{tab:cent_mpl} \vspace{-.5em}
\end{table}

\noindent\textbf{Federated UDA.}
For the federated UDA tasks, we follow the procedure detailed in \cref{alg:fed_mapseg}. We initialize the encoder of the global model $f_\phi$ with the encoder pretrained on the large-scale data mentioned in Sec.4.3. We set the global FL round $R = 100$. We set both the server and client update steps to 1 epoch with batch size of 1. Training configuration inherits mostly from that of the centralized UDA, except a global cosine annealing learning rate schedule is adopted to decay the learning rate from $1e^{-4}$ to $1e^{-6}$ over the course of the FL rounds.

\noindent\textbf{Test-Time UDA.}
For the test-time UDA tasks, we follow the same configuration as listed in \hyperref[tab:cent_mpl]{Suppl.Tab.3}. The difference is that the model can only access source domain data (image and label) during warmup epochs and can only access target domain data (image only) after that, while centralized UDA has synchronous access to both source and target domain data throughout the whole training process.
\begin{algorithm}[tb]
\caption{Federated MAPSeg (Fed-MAPSeg)}
\label{alg:fed_mapseg}
\begin{algorithmic}[1]
\REQUIRE Source domain dataset $D_S = \{(x_s, y_s)\}$ and target domain datasets $D_T^k = \{(x_t^k)\}$ for each client $k$, pretrained global model $f_\phi$, number of FL round $R$, number of server update steps $T_s$, number of client update steps $T_t$
\FOR{$r=1,2,\cdots, R$}
    \STATE Initialize server EMA teacher model: $\theta \gets \phi$
    \FOR{$t=1,2,\cdot,T_s$}
        \STATE Sample patches $(x_s, y_s)$ from $D_S$ and generate downsampled global volume and masked inputs $X_s$, $X_s^M$, $x_s^M$
        \STATE Update $f_\phi$ on server by minimizing $\mathcal{L}_s$ (Eq.9)
        \STATE Update server EMA teacher model parameter $\theta$ with (Eq.3)
    \ENDFOR
    \STATE Server broadcast $\theta$ to clients
    \FOR{each client $k$ in parallel}
        \STATE $\phi_k \gets \theta$, $\theta_k \gets \theta$
        \FOR{$t=1,2,\cdots,T_t$}
            \STATE Sample patches $x_t^k$ from $D_T^k$ and generate downsampled global volume and masked inputs $X_t^k$, $(X_t^k)^M$, $(x_s^k)^M$
            \STATE Generate pseudolabels for unmasked inputs $x_t^k$ and $X_t^k$ using the teacher model $f_{\theta_k}$: $f_{\theta_k}(x_t^k)$ and $f_{\theta_k}(X_t^k)$
            \STATE Update $f_{\phi_k}$ by minimizing $\mathcal{L}_u$ (Eq.10)
            \STATE Update client EMA teacher model parameter with (Eq.3)
        \ENDFOR
        \STATE Upload $\theta_k$ to server
    \ENDFOR
\STATE The server aggregates $\theta_k$ from clients:
\[\Bar{\theta} \gets \sum_k \frac{|D_T^k|}{\sum_k |D_T^k|}\theta_k\]
\STATE Update server model parameters $\phi \gets \Bar{\theta}$
\ENDFOR
\end{algorithmic}
\end{algorithm}
\subsection{Implementation of Comparing Methods}
\label{sec:baseline}
For other comparing methods in centralized UDA, we adapt their official implementations. For DAFormer, HRDA, and MIC, we modify the ground truth labels to make them denser, as we observe that the original sparse annotations cause trouble for those methods. Specifically, we crop the scans to include only brain regions. In addition to having foreground classes of 7 subcortical regions (which account for approximately 2\% of overall voxels), we assign another foreground class to the remaining brain regions. Therefore, there are 9 classes for DAFormer, HRDA, and MIC, 8 foreground and 1 background classes. This modification significantly improves the results. For the FL baselines FAT~\cite{DBLP:conf/isbi/MushtaqBDA23} and DualAdapt~\cite{fmtda}, since there is no public official implementation available, we implement both methods following the description in the original papers and finetune thoroughly. We use the same network backbone initialized with the same pretrained encoder and training configuration (FL rounds, global learning rate schedule, local update steps, batch size, etc.) as Fed-MAPSeg whenever possible.

\subsection{Dataset Description}
\label{sec:datadetail}
We include a diverse collection of 2,421 brain MRI scans from several international projects, each with its unique focus on infant brain development. From the Developing Human Connectome Project (dHCP) V1.0.2 data release\footnote{\url{https://www.developingconnectome.org/data-release/data-release-user-guide/}} \cite{edwards2022developing} in the UK, we incorporate 983 scans (426 T1-weighted, T1w), acquired shortly after birth. The Baby Connectome Project (BCP) \cite{howell2019unc} in the USA contributes 892 scans (519 T1w), featuring longitudinal data. Additionally, from the Environmental Influences on Child Health Outcomes (ECHO) project, also in the USA, we have 433 scans (218 T1w) from newborn infants. The ‘Maternal Adversity, Inflammation, and Neurodevelopment’ (Healthy Minds) project from Brazil, conducted at Hospital São Paulo - Federal University of São Paulo (UNIFESP), adds 103 T2-weighted (T2w) MRI scans, acquired shortly after birth and available in the National Institute of Mental Health Data Archive (collection ID 3811). Lastly, the Melbourne Children’s Regional Infant Brain (M-CRIB) project \cite{mcrib} from Australia provides 10 additional T2w scans. All studies involved have received Institutional Review Board (IRB) approvals. MAPSeg takes normalized scans as inputs. During training, the intensity of each volumetric scan is clipped at a percentile randomly drawn from a uniform distribution $\mathcal{U}(99,100)$, then normalized to 0-1. During inference, the intensity clip is fixed at 99.5\%. The top 0.5\% intensity is clipped as 1 to cope with outlier pixels (hyperintensities) that are usual in MRI. 

\subsection{Results of MRI $\rightarrow$ CT cardiac segmentation}
\label{sec:mri2ctresult}
The performance of MAPSeg on the public cardiac MRI$\rightarrow$CT segmentation is reported in \hyperref[tab:mri2ct]{Suppl.Tab.4}. Similarly, we use the same dataset partition as previous studies. MAPSeg consistently outperforms other baseline methods, although the performance gap is smaller than CT$\rightarrow$MRI.
\begin{table}
  \caption{Results of cardiact MRI$\rightarrow$CT segmentation.}
  \vspace{-1em}
  \centering
  \begin{center}
      \resizebox{0.78\linewidth}{!}{%
          \begin{tabular}{c|cccccccc}
              \toprule
              \multicolumn{6}{c}{Cardiac CT $\rightarrow$ MRI segmentation}\\	
              \hline
              \multirow{2}{*}{Method} &\multicolumn{5}{c}{Dice(\%) $\uparrow$} \\
              \cline{2-6}
              &AA &LAC &LVC &MYO &Avg \\
              \hline
              PnP-AdaNet\cite{8764342} &74.0&68.9&61.9&50.8&63.9 \\
              SIFA-V1\cite{Chen_Dou_Chen_Qin_Heng_2019} &81.1&76.4&75.7&58.7&73.0 \\
              SIFA-V2\cite{8988158} &81.3&79.5&73.8&61.6&74.1 \\
              DAFormer\cite{daformer} &85.5&88.2&74.5&60.2&77.1 \\
              MPSCL\cite{9672690} & 90.3&87.1&86.5&72.5&84.1\\
              MA-UDA\cite{10273225} &90.8&88.7&77.6&67.4&81.1 \\
              SE-ASA\cite{Feng_Ju_Wang_Song_Zhao_Ge_2023} &83.8&85.2&82.9&71.7&80.9 \\
              FSUDA-V1\cite{Liu_Yin_Qu_Wang_2023} &86.4&86.9&84.8&81.8&85.0\\
              PUFT\cite{10021602} &88.1&88.5&87.5&74.1&84.6\\
              SDUDA\cite{Cui_structure_driven}&87.9&88.1&88.4&78.7&85.8\\
              FSUDA-V2\cite{10261458} &88.2&\textbf{88.9}&85.2&\textbf{82.2}&86.1\\
              
              \hline
              \textbf{MAPSeg (Ours)}&\underline{\textbf{93.3}}&\underline{87.3}&\underline{\textbf{89.1}}&\underline{78.9}&\underline{\textbf{87.1}}\\

              \bottomrule

      \end{tabular}}
  \end{center}
  \label{tab:mri2ct}
  \vspace{-1.5em}
\end{table}

\subsection{Additional Analysis}
\label{sec:additionres}
\begin{figure}
  \centering
     \includegraphics[width=1\linewidth]{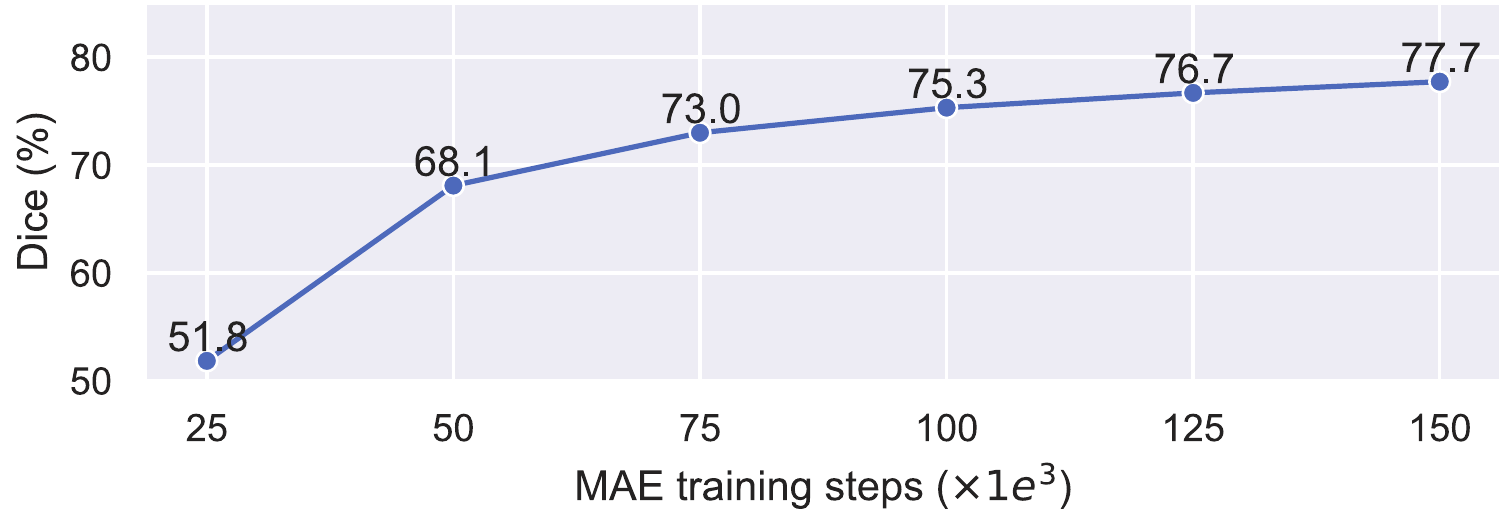}
  \caption{Downstream cross-sequence centralized UDA performance vs. MAE pretraining iterations.}
  \label{fig:maesteps}
\end{figure}

\noindent\textbf{Influence of MAE Pretraining on UDA Results.} We conduct an additional analysis to investigate the relationship between MAE training steps and downstream UDA performance. The experiments are conducted on cross-sequence brain MRI segmentation (\hyperref[fig:maesteps]{Suppl.Fig.2}). We observe significant improvement in UDA performance at the first 75,000 MAE training steps, which then gradually saturates. We choose 150,000 MAE training steps as the benefits of further training diminish. 

\noindent\textbf{Sensitivity to hyperparameters.} We conduct additional experiments on cross-sequence brain MRI segmentation to investigate the sensitivity of MAPSeg to hyperparameters (\hyperref[tab:hyperparams]{Suppl.Tab.5}). Specifically, we investigate the step size ($\alpha$) of EMA update as well as weights of loss terms ($\gamma$ and $\delta$). When one parameter is varying, other parameters remain unchanged. We notice that the performance is relatively stable across a wide range of hyperparameters. Since we did not tune the hyperparameters extensively during development, the default parameters may not represent the optimal setting.

\begin{table}[h]
    \caption{Influence of hyperparameters on results, bold indicates used parameters.}
    \centering
    \label{tab:hyperparams}
    \begin{center}
        \resizebox{0.77\linewidth}{!}{%
            \begin{tabular}{c|ccccc}
                \hline

                $\alpha$ &\textbf{0.999/0.9999} &0.99/0.999 &0.99 &0.999 &0.9999 \\
                \hline
                Dice (\%) &77.73&74.00&74.26&74.74&78.06 \\
                \hline

                $\gamma$ &\textbf{0.05} & 0.5 & 0.1 & 0.01 & 0.005  \\
                \hline
                Dice (\%) &77.73 & 77.22 & 77.97 & 77.98 & 77.99 \\
                \hline

                $\delta$ &\textbf{0.025} &0.25 &0.1 &0.01 &0.0025 \\
                \hline
                Dice (\%) &77.73 & 76.74 & 78.08 & 77.82 & 78.57 \\
                \hline
        \end{tabular}}
    \end{center}
\end{table}

\subsection{Visualization}
\label{sec:appendixvis}
\begin{figure}
  \centering
     \includegraphics[width=1\linewidth]{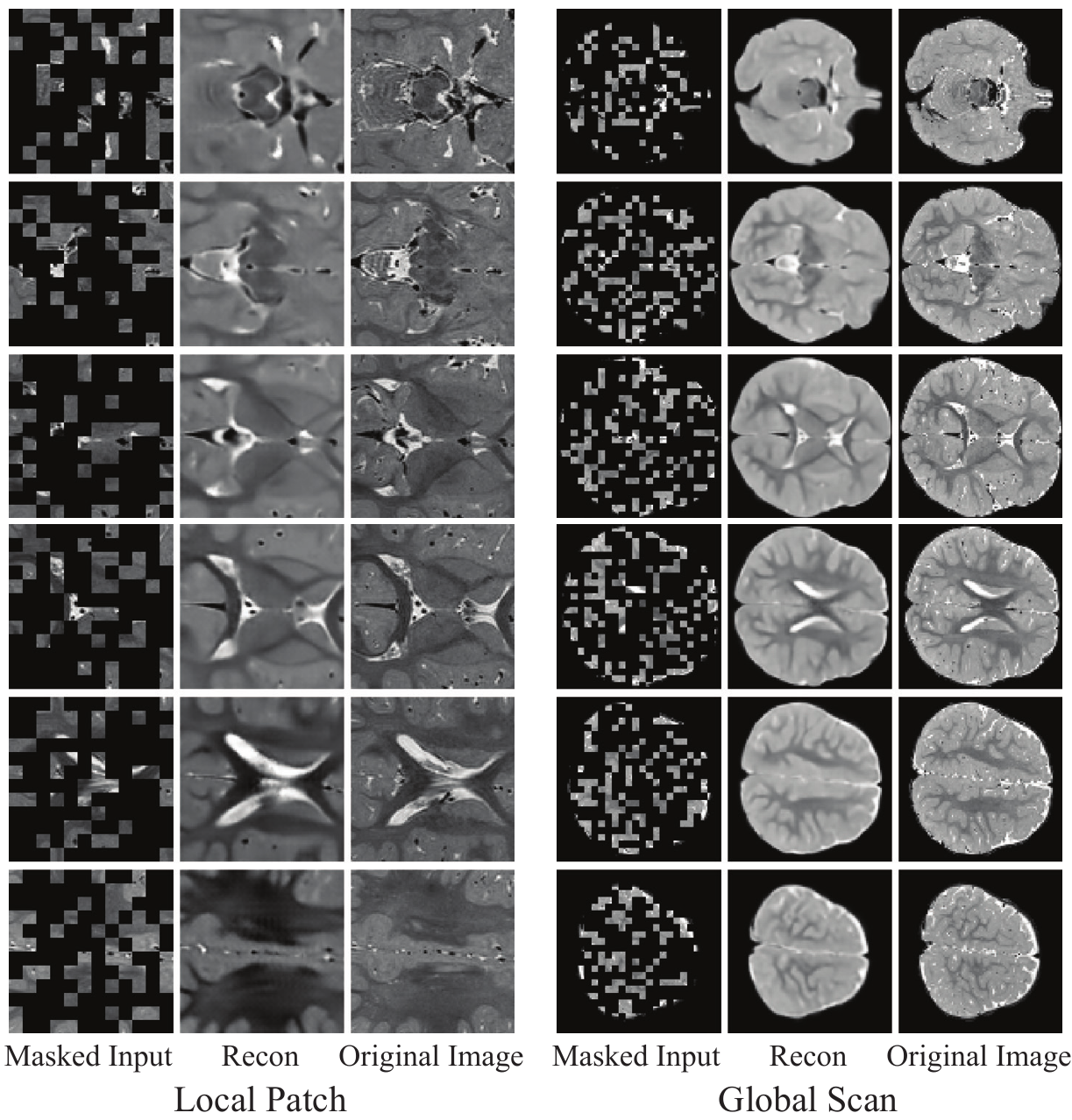}
  \caption{A randomly sampled T2w scan in cross-sequence task. MAE parameters is same as in \hyperref[tab:mae_pretrain]{Suppl.Tab.2}}
  \label{fig:maeres}
\end{figure}

\textbf{MAE.}
Some visualizations of MAE results (axial slices) are provided in \hyperref[fig:maeres]{Suppl.Fig.3}. 

\begin{figure*}
  \centering
     \includegraphics[width=1\linewidth]{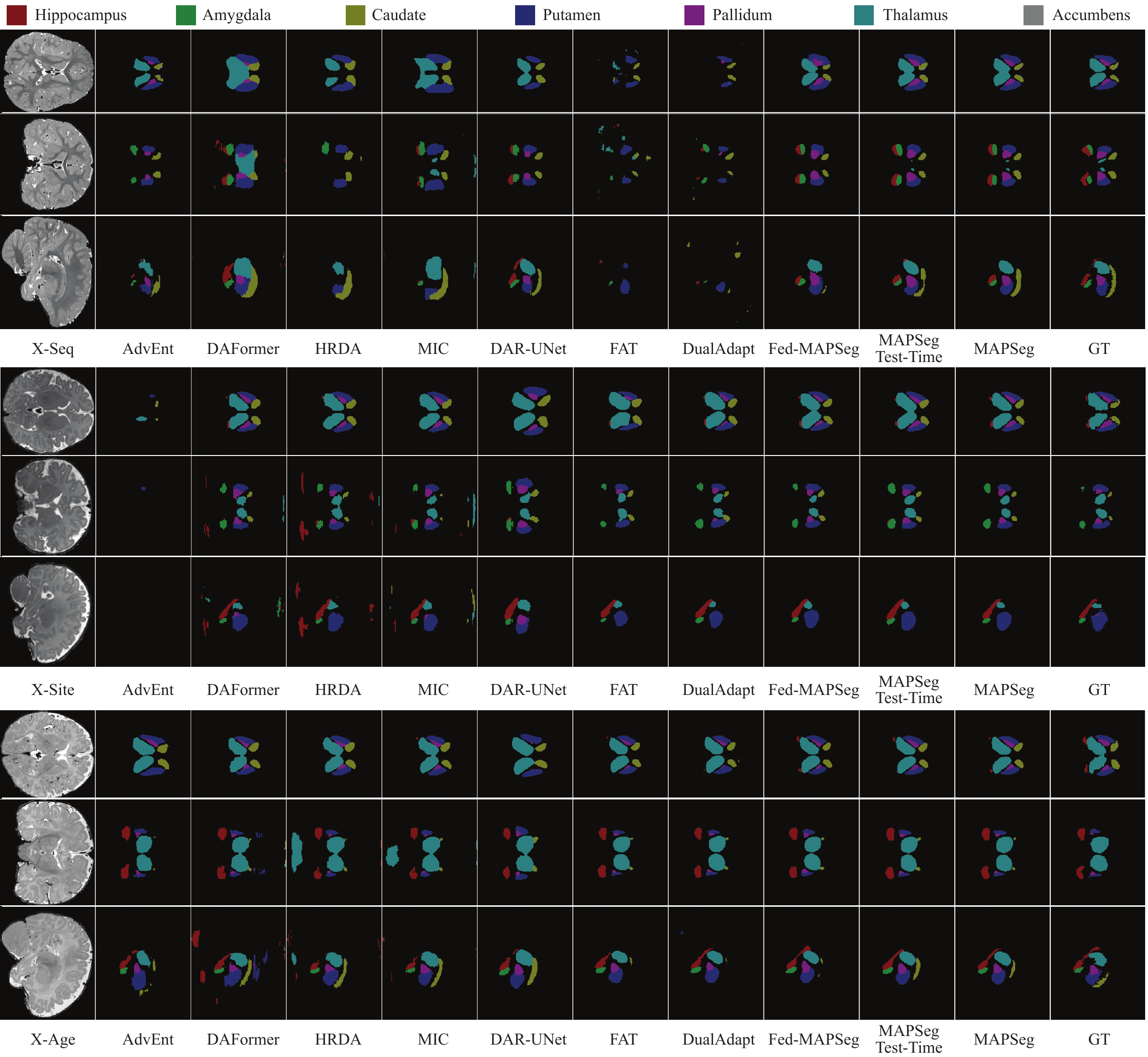}
  \caption{Qualitative comparisons. Three rows (top to bottom) of each task represent axial plane, coronal plane, and sagittal plane, respectively.}
  \label{fig:visres}
\end{figure*}

\noindent\textbf{UDA Results.}
We provide qualitative comparisons of different methods on cross-sequence (X-Seq), cross-site (X-Site), and cross-age (X-Age) brain MRI segmentation tasks in \hyperref[fig:visres]{Suppl.Fig.4}. MAPSeg consistently provides accurate segmentation in different UDA settings. It is worth noting that, despite the second best performance in cross-sequence, DAR-UNet tends to oversegment on cross-site and cross-age tasks, partially because of translation errors. On cross-site and cross-age tasks, despite DAFormer, HRDA, and MIC generate reasonably good segmentation inside the subcortical regions, they exhibit extensive false positives outside the subcortical regions, leading to suboptimal overall Dice score.

\end{document}